\documentclass[runningheads]{llncs}
\usepackage{graphicx}

\usepackage{tikz}
\usepackage{comment}
\usepackage{color}

\usepackage[utf8]{inputenc} %
\usepackage[T1]{fontenc}    %
\usepackage{hyperref}       %
\usepackage{url}            %
\usepackage{booktabs}       %
\usepackage{amsfonts}       %
\usepackage{nicefrac}       %
\usepackage{microtype}      %
\usepackage{xcolor}         %
\usepackage{amssymb}
\usepackage{amsmath}
\usepackage{dsfont}
\usepackage{algorithmic}
\usepackage[figuresright]{rotating}
\usepackage{multirow}
\definecolor{Gray}{gray}{0.9}
\usepackage{colortbl}
\usepackage{wrapfig}
\usepackage{blindtext}
\usepackage{listings}
\newcommand{\methodAbbr}{DICE~}

\def\*#1{\mathbf{#1}}

\definecolor{codegreen}{rgb}{0,0.6,0}
\definecolor{codegray}{rgb}{0.5,0.5,0.5}
\definecolor{codepurple}{rgb}{0.58,0,0.82}
\definecolor{backcolour}{rgb}{0.95,0.95,0.92}

\lstdefinestyle{mystyle}{
    backgroundcolor=\color{backcolour},   
    commentstyle=\color{codegreen},
    keywordstyle=\color{magenta},
    numberstyle=\tiny\color{codegray},
    stringstyle=\color{codepurple},
    basicstyle=\ttfamily\footnotesize,
    breakatwhitespace=false,         
    breaklines=true,                 
    captionpos=b,                    
    keepspaces=true,                 
    numbers=left,                    
    numbersep=5pt,                  
    showspaces=false,                
    showstringspaces=false,
    showtabs=false,                  
    tabsize=2
}

\usepackage[accsupp]{axessibility}  

\begin{document}
\pagestyle{headings}
\mainmatter
\def\ECCVSubNumber{4405}  

\title{DICE: Leveraging Sparsification for Out-of-Distribution Detection}

\titlerunning{DICE: Leveraging Sparsification for Out-of-Distribution Detection}

\author{Yiyou Sun \and
Yixuan Li}

\authorrunning{Y. Sun, Y. Li}
\institute{Computer Science Department, University of Wisconsin-Madison \\
\email{\{sunyiyou, sharonli\}@cs.wisc.edu}}

\maketitle

\begin{abstract}
Detecting out-of-distribution (OOD) inputs is a central challenge for safely deploying machine learning models in the real world. Previous methods commonly rely on an OOD score derived from the overparameterized weight space, while largely overlooking the role of \emph{sparsification}. In this paper, we reveal important insights that reliance on unimportant weights and units can directly attribute to the brittleness of OOD detection. To mitigate the issue, we propose a sparsification-based OOD detection framework termed \textbf{DICE}. Our key idea is to rank weights based on a measure of contribution, and selectively use the most salient weights to derive the output for OOD detection. We provide both empirical and theoretical insights, characterizing and explaining the mechanism by which DICE improves OOD detection. By pruning away noisy signals, DICE provably reduces the output variance for OOD data, resulting in a sharper output distribution and stronger separability from ID data. We demonstrate the effectiveness of
sparsification-based OOD detection on several
benchmarks and establish competitive performance.
Code is available at:  \url{https://github.com/deeplearning-wisc/dice.git}.

\keywords{Out-of-distribution Detection, Sparsification}
\end{abstract}


\section{Introduction}
\label{sec:intro}
Deep neural networks deployed in real-world systems often encounter out-of-distribution (OOD) inputs---samples from unknown classes that the network has not been exposed to during training, and therefore should not be predicted by the model in testing. 
Being able to estimate and mitigate OOD uncertainty is paramount for safety-critical applications such as medical diagnosis~\cite{roy2021does,wang2017chestx} and autonomous driving~\cite{filos2020can}. For example, an autonomous vehicle may fail to recognize objects on the road that do not appear in its detection model’s training set, potentially leading to a crash. This gives rise to the importance of OOD detection, which allows the
learner to express ignorance and take precautions in the presence of OOD data.  %

The main challenge in OOD detection stems from the fact that modern deep neural networks can easily produce overconfident predictions on OOD inputs, making the separation between in-distribution (ID) and OOD data a non-trivial task. The vulnerability of machine learning to OOD data can be hard-wired in high-capacity models used in practice. In particular, modern deep neural networks can overfit observed patterns in the training data~\cite{zhang2016understanding}, and worse, {activate features on unfamiliar inputs}~\cite{nguyen2015deep}. 
To date, existing OOD detection methods commonly derive OOD scores using overparameterized weights, while largely overlooking the role of \emph{sparsification}. This paper aims to bridge the gap.

\begin{figure}[t]
	\begin{center}
		\includegraphics[width=0.9\linewidth]{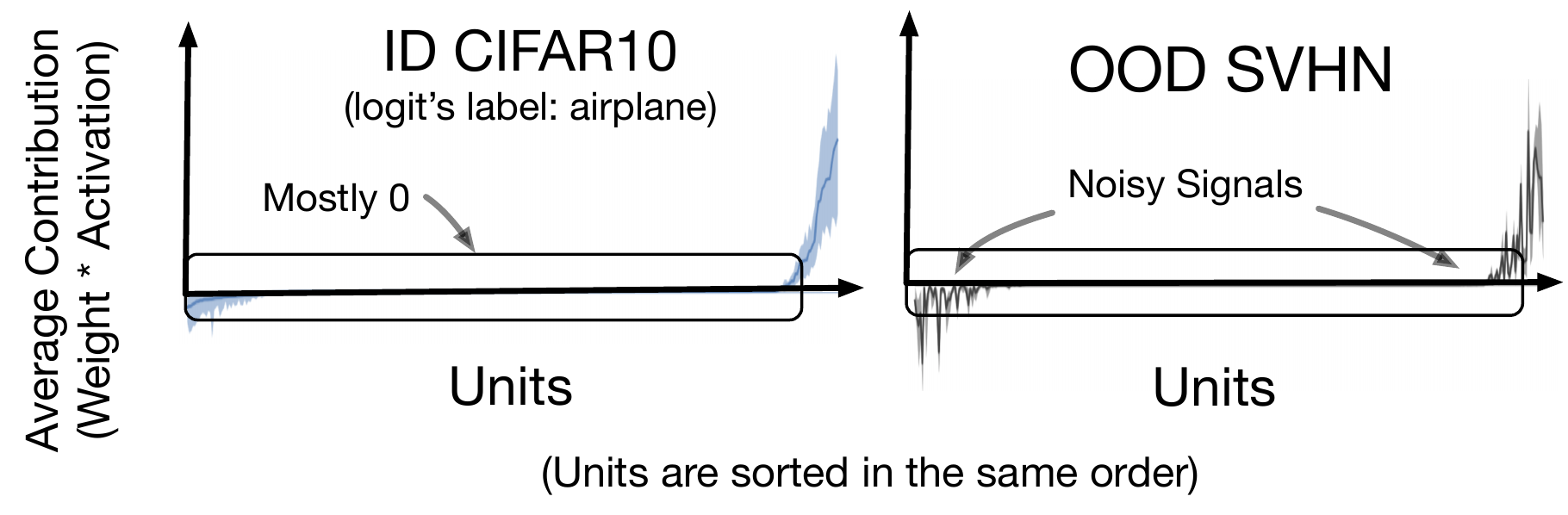}
	\end{center}
	\caption{\small Illustration of unit contribution (\emph{i.e.}, \texttt{weight} $\times$ \texttt{activation}) to the class output. For class $c$, the output $f_c(\*x)$ is the summation of unit contribution from the penultimate feature layer of a neural network. \emph{Units are sorted in the same order}, based on the expectation of ID data's contribution (averaged over many CIFAR-10 samples) on the $x$-axis. \textbf{Shades indicate the variance for each unit}. \textbf{Left:} For in-distribution data (CIFAR-10, airplane), only a subset of units contributes to the model output.  \textbf{Right:} In contrast, out-of-distribution (OOD) data can trigger a non-negligible fraction of units with noisy signals, as indicated by the variances.}
	\label{fig:whytopk}
\end{figure}

In this paper, we start by revealing key insights that reliance on unimportant units and weights can directly attribute to the brittleness of OOD detection. Empirically on a network trained with CIFAR-10, we show that an OOD image can activate a non-negligible fraction of units in the penultimate layer (see Figure~\ref{fig:whytopk}, right). Each point on the horizontal axis corresponds to a single unit. The y-axis measures the unit contribution (\emph{i.e.}, \texttt{weight} $\times$ \texttt{activation}) to the output of class \textsc{airplane}, with the solid line and the shaded area indicating the mean and variance, respectively. 
Noticeably, for OOD data (gray), we observe a non-negligible fraction of ``noisy'' units that display high variances of contribution, which is then aggregated to the model's output through summation. As a result, such noisy signals can undesirably manifest in model output---increasing the variance of output distribution and reducing the separability from ID data.

The above observation motivates a simple and effective method,
\emph{\textbf{Di}rected \textbf{S}parisification} (\textbf{DICE}), for OOD detection. %
\methodAbbr leverages the observation that a model's prediction for an ID class depends
on only a subset of important units (and corresponding weights), as evidenced in Figure~\ref{fig:whytopk} (left). To exploit this, our novel idea is to rank weights based on the measure of {contribution}, and selectively use the most contributing weights to derive the output for OOD detection. As a result of the weight sparsification, we show that the model's output becomes more separable between ID and OOD data.   Importantly, DICE can be conveniently used by {{post hoc} weight masking} on a pre-trained network and therefore can preserve the ID classification accuracy. 
Orthogonal to existing works on sparsification for accelerating computation, our primary goal is to explore the sparsification approach for improved  OOD detection performance.%

\begin{figure*}[tb]
	\begin{center}
		\includegraphics[width=0.95\linewidth]{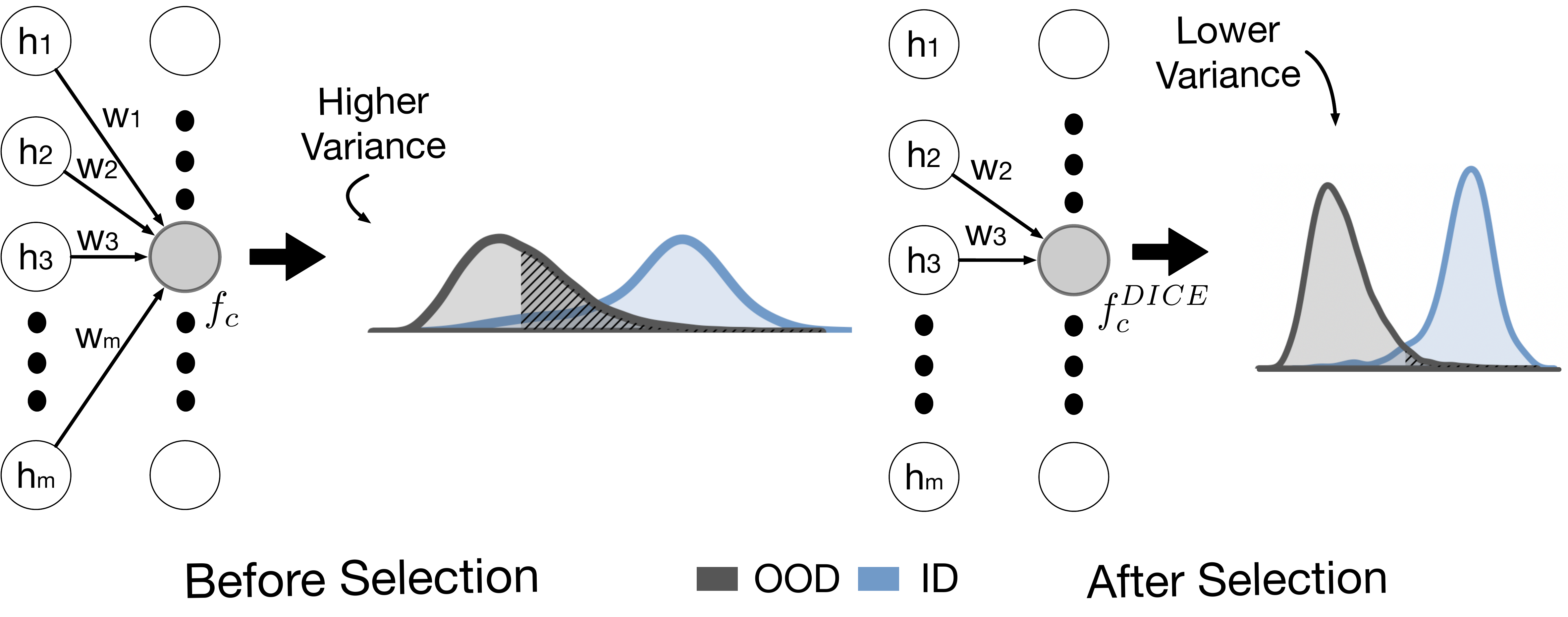}
	\end{center}
	\caption{\small Illustration of out-of-distribution detection using \emph{Directed Sparsification} (\textbf{DICE}). We consider a pre-trained neural network, which encodes an input $\*x$ to a feature vector $h(\*x) \in \mathbb{R}^m$. \textbf{Left}: The logit output $f_c(\*x)$ of class $c$ is a linear combination of activation from \emph{all} units in the preceding layer, weighted by $w_i$. The full connection results in a high variance for OOD data's output, as depicted in the gray. \textbf{Right}: Our proposed approach leverages a selective subset of weights, which effectively reduces the output variance for OOD data, resulting in a sharper score distribution and stronger separability from ID data. The output distributions are based on CIFAR-10 trained network, with ID class label ``frog'' and SVHN as OOD.}
	\label{fig:teaser}
\end{figure*}

We provide both empirical and theoretical insights characterizing and explaining the mechanism by which \methodAbbr improves OOD detection. We perform extensive evaluations and establish competitive performance on common OOD detection benchmarks, including CIFAR-10, CIFAR-100~\cite{krizhevsky2009learning}, and a large-scale ImageNet benchmark~\cite{huang2021mos}. Compared to the competitive post hoc method ReAct~\cite{sun2021react}, \methodAbbr reduces the FPR95 by up to {12.55}\%. Moreover, we perform ablation using various sparsification techniques and demonstrate the benefit of {directed sparsification} for OOD detection. 
Theoretically, by pruning away noisy signals from unimportant units and weights,
\methodAbbr  \emph{provably reduces the output variance}  and results in a sharper output distribution (see Section~\ref{sec:theory}). The sharper distributions lead to a stronger separability between ID and OOD data and overall improved OOD detection performance (\emph{c.f.} Figure~\ref{fig:teaser}).  Our \textbf{key results and contributions} are:
\begin{itemize}
    \item (Methodology) We introduce DICE, a simple and effective approach for OOD detection utilizing {post hoc} weight sparsification. To the best of our knowledge, DICE is the first to explore and demonstrate the effectiveness of sparsification for OOD detection.  
    
    \item (Experiments) We extensively evaluate \methodAbbr on common benchmarks and establish competitive performance among post hoc OOD detection baselines. \methodAbbr  outperforms the strong baseline~\cite{sun2021react} by reducing the FPR95 by up to {12.55}\%. We show \methodAbbr can effectively improve OOD detection while preserving the classification accuracy on ID data.
    
    \item (Theory and ablations) We provide  ablation and theoretical analysis that improves understanding of a sparsification-based method for OOD detection. Our analysis reveals an important variance reduction effect, which provably explains the effectiveness of DICE. We hope our insights inspire future research on weight sparsification for OOD detection. 
\end{itemize}


\section{Preliminaries}
\label{sec:background}

We start by recalling the general setting of the supervised learning problem. We denote by $\mathcal{X}=\mathbb{R}^d$  the input space and $\mathcal{Y}=\{1,2,...,C\}$  the output space. A learner is given access to a set of training data $\mathcal{D}=\{(\*x_i,y_i)\}_{i=1}^N$ drawn from an unknown joint data distribution $\mathcal{P}$ defined on $\mathcal{X}\times \mathcal{Y}$. Furthermore, let $\mathcal{P}_\text{in}$ denote the marginal probability distribution on $\mathcal{X}$. \\

\noindent \textbf{Out-of-distribution detection} When deploying a model in the real world, a reliable classifier should not only accurately classify known in-distribution (ID) samples, but also identify any OOD input as ``unknown''. This can be achieved through having dual objectives: ID/OOD classification and multi-class classification of ID data~\cite{bendale2016towards}. %

OOD detection can be formulated as a binary classification problem.  At test time, the goal of OOD detection is to decide whether a sample $\*x \in \mathcal{X}$ is from $\mathcal{P}_\text{in}$ (ID) or not (OOD). In literature, OOD distribution $\mathcal{P}_\text{out}$ often simulates unknowns encountered during deployment time, such as samples from an irrelevant distribution {whose label set has no intersection with $\mathcal{Y}$ and therefore should not be predicted by the model}. The decision can be made via a thresholding comparison: %
\begin{align*}
\label{eq:threshold}
	g_{\lambda}(\*x)=\begin{cases} 
      \text{in} & S(\*x)\ge \lambda \\
      \text{out} & S(\*x) < \lambda 
  \end{cases},
\end{align*}
where samples with higher scores $S(\*x)$ are classified as ID and vice versa, and  $\lambda$ is the threshold.

\section{Method}
\label{sec:method}

\noindent \textbf{Method overview} %
Our novel idea is to selectively use a subset of important weights to derive the output for OOD detection. By utilizing sparsification, the network prevents adding irrelevant information to the output. We illustrate our idea in Figure~\ref{fig:teaser}. Without DICE (\emph{left}), the final output is a summation of weighted activations across all units, which can have a high variance for OOD data (colored in gray). In contrast, with DICE (\emph{right}), the variance of output can be significantly reduced, which improves separability from ID data. We proceed with describing our method in details, and provide the theoretical explanation later in Section~\ref{sec:theory}.

\subsection{DICE: Directed Sparsification}
\label{sec:dks}
We consider a deep neural network parameterized by $\theta$, which encodes an input $\*x \in \mathbb{R}^d$ to a feature space with dimension $m$. We denote by $ h(\*x) \in \mathbb{R}^m$ the feature vector from the penultimate layer of the network.
A weight matrix $\mathbf{W} \in \mathbb{R}^{m\times C}$ connects the feature $h(\*x)$ to the output $f(\*x)$. \\

\noindent \textbf{Contribution matrix} We perform a \emph{directed sparsification} based on a measure of contribution, and preserve the most important weights in $\*W$. To measure the contribution, we define a contribution matrix $\*V \in \mathbb{R}^{m\times C}$, where each column $\*v_c \in \mathbb{R}^m$ is given by:
\begin{align}
\*v_c = \mathbb{E}_{\*x\in \mathcal{D}} [\*w_c \odot h(\*x)],
\end{align}
where $\odot$ indicates the element-wise multiplication, and $\*w_c$ indicates weight vector for class $c$. Each element in $\*v_c \in \mathbb{R}^m$ intuitively measures the corresponding unit's average contribution to class $c$, estimated empirically on in-distribution data $\mathcal{D}$. A larger value indicates a higher contribution to the output $f_c(\*x)$ of class $c$. The vector $\*v_c$ is derived for all classes  $c \in \{1,2,...,C\}$, forming the contribution matrix $\*V$. Each element $\*v_{c}^i\in\*V$ measures the average contribution (\texttt{weight} $\times$ \texttt{activation}) from a unit $i$ to the output class $c\in \{1,2,...,C\}$. %

We can now select the top-$k$ weights based on the $k$-largest elements in $\*V$.  In particular, we define a masking matrix $\*M \in \mathbb{R}^{m\times C}$, which returns a matrix by setting $1$ for entries corresponding to the $k$ largest elements in $\*V$ and setting other elements to $0$.
The model output under \emph{contribution-directed sparsification} is given by

\begin{align}
f^{\text{DICE}}(\*x;\theta) = (\*M\odot\*W)^\top h(\*x) + \mathbf{b},
\end{align}
where $\mathbf{b} \in \mathbb{R}^C$ is the bias vector. The procedure described above essentially accounts for information from the most relevant units in the penultimate layer. Importantly, the sparsification can be conveniently imposed by \emph{post hoc} weight masking on the final layer of a pre-trained network, without changing any parameterizing of the neural network. Therefore one can improve OOD detection while preserving the ID classification accuracy. \\

\noindent \textbf{Sparsity parameter $p$} To align with the convention in literature, we use the sparsity parameter $p= 1- \frac{k}{m\cdot C}$ in the remainder paper. A higher $p$ indicates a larger fraction of weights dropped. When $p=0$, the output becomes equivalent to the original output $f(\*x;\theta)$ using dense transformation, where $f(\*x;\theta) = \*W^\top h(\*x) + \*b$. We provide ablations on the sparsity parameter later in Section~\ref{sec:sparsification}.
\subsection{OOD Detection with DICE}
Our method \methodAbbr  in Section~\ref{sec:dks} can be flexibly leveraged by the downstream OOD scoring function:
\begin{align}
\label{eq:threshold_dice}
	g_{\lambda}(\*x)=\begin{cases} 
      \text{in} & S_\theta(\*x)\ge \lambda \\
      \text{out} & S_\theta(\*x) < \lambda
   \end{cases},
\end{align}
 where a thresholding mechanism is exercised to distinguish between ID and OOD during test time. The threshold $\lambda$ is typically chosen so that a high fraction of ID data (\emph{e.g.,} 95\%) is correctly classified. Following recent work by Liu \emph{et. al}~\cite{liu2020energy}, we derive an energy score using the logit output $f^\text{DICE}(\*x;\theta)$ with contribution-directed sparsification. The function maps the logit outputs $f^\text{DICE}(\*x;\theta)$ to a scalar $E_\theta(\*x) \in \mathbb{R}$, which is relatively lower for ID data:
 \begin{equation}
     S_\theta(\*x) = -E_\theta(\*x) = \log \sum_{c=1}^C \exp(f_c^\text{DICE}(\*x;\theta)).
 \end{equation}
 The energy score can be viewed as the log of the denominator in softmax function:
 \begin{equation}
     p(y | \*x) = \frac{p(\*x,y)}{p(\*x)} = \frac{\exp(f_y(\*x;\theta))}{\sum_{c=1}^C \exp(f_c(\*x;\theta))},
 \end{equation}
and enjoys better theoretical interpretation than using posterior probability $p(y|\*x)$. 
Note that DICE can also be compatible with an alternative scoring function such as maximum softmax probability (MSP)~\cite{Kevin}, though the performance of MSP is less competitive (see Appendix~\ref{sec:dice_msp}). Later in Section~\ref{sec:theory}, we formally characterize and explain why DICE improves the separability of the  scores between ID and OOD data. 
\label{sec:ood}


\section{Experiments}
\label{sec:experiments}
In this section, we evaluate our method on a suite of OOD detection tasks. We begin with the CIFAR benchmarks that are routinely used in literature (Section~\ref{sec:common_benchmark}).  In Section~\ref{sec:imagenet}, we continue with a large-scale OOD detection task based on ImageNet. 
\subsection{Evaluation on Common Benchmarks}
\label{sec:common_benchmark}

\noindent \textbf{Experimental details}
We use  CIFAR-10~\cite{krizhevsky2009learning}, and CIFAR-100~\cite{krizhevsky2009learning} datasets as in-distribution data. We use the standard split with 50,000 training images and 10,000 test images. We evaluate the model on six common OOD benchmark datasets: \texttt{Textures}~\cite{cimpoi2014describing}, \texttt{SVHN}~\cite{netzer2011reading}, \texttt{Places365}~\cite{zhou2017places}, \texttt{LSUN-Crop}~\cite{yu2015lsun}, \texttt{LSUN-Resize}~\cite{yu2015lsun}, and \texttt{iSUN}~\cite{xu2015turkergaze}. We use DenseNet-101 architecture~\cite{huang2017densely} and train on in-distribution datasets. The feature dimension of the penultimate layer is 342. For both CIFAR-10 and CIFAR-100, the model is trained for 100 epochs with batch size 64, weight
decay 0.0001 and momentum 0.9. The start learning rate is 0.1 and decays by a factor of 10 at epochs 50, 75, and 90. We use the validation strategy in Appendix~\ref{app:val} to select $p$. \\

\noindent \textbf{DICE vs. competitive baselines} We show the results in Table~\ref{tab:cifar-results}, where \methodAbbr outperforms competitive baselines. In particular, we compare with Maximum Softmax Probability~\cite{Kevin}, ODIN~\cite{liang2018enhancing}, Mahalanobis distance~\cite{lee2018simple}, Generalized ODIN~\cite{godin2020CVPR}, Energy score~\cite{liu2020energy}, and ReAct~\cite{sun2021react}. For a fair comparison, all the methods derive the OOD score post hoc from the same pre-trained model, except for G-ODIN which requires model re-training. {For readers' convenience, a brief introduction of baselines and hyperparameters is provided in Appendix~\ref{sec:baseline}.} 

On CIFAR-100, we show that \methodAbbr reduces the average FPR95 by \textbf{18.73\%} compared to the vanilla energy score~\cite{liu2020energy} without sparsification. Moreover, our method also outperforms a competitive method ReAct~\cite{sun2021react} by {12.55\%}. While {ReAct} only considers activation space, {DICE} examines \emph{both the {weights} and activation} values together---the multiplication of which {directly} determines the network's logit output. 
{Overall our method is more generally applicable}, and can be implemented through a simple {post hoc} weight masking. \\

\begin{table}[t]
\centering
\caption{\small Comparison with competitive \textit{post hoc} out-of-distribution detection method on CIFAR benchmarks. All values are percentages and are averaged over 6 OOD test datasets. The full results for each evaluation dataset are provided in Appendix~\ref{sec:detailed-cifar}. We report standard deviations estimated across 5 independent runs. $^\S$ indicates an exception, where model retraining using a different loss function is required.} 
\scalebox{0.9}{
\begin{tabular}{lll|lll}
\toprule
 \multicolumn{1}{c}{\multirow{3}{*}{\textbf{Method}}} & \multicolumn{2}{c}{\textbf{CIFAR-10}} & \multicolumn{2}{c}{\textbf{CIFAR-100}} \\
 \multicolumn{1}{c}{} & \multicolumn{1}{l}{\textbf{FPR95}} & \multicolumn{1}{l|}{\textbf{AUROC}} & \multicolumn{1}{l}{\textbf{FPR95}} & \multicolumn{1}{l}{\textbf{AUROC}} & \\ 
  & $\downarrow$ & $\uparrow$ & $\downarrow$ & $\uparrow$ & \\ \midrule
 MSP~\cite{Kevin}  & 48.73 & 92.46 & 80.13 & 74.36 \\
 ODIN~\cite{liang2018enhancing}  & 24.57 & 93.71 & 58.14 & 84.49  \\
  GODIN$^\S$~\cite{godin2020CVPR}  & 34.25 & 90.61 & 52.87 & 85.24  \\
 Mahalanobis~\cite{lee2018simple}  & 31.42	 & 89.15 & 55.37 &	82.73 \\
 Energy~\cite{liu2020energy}  & 26.55 & 94.57 & 68.45 & 81.19  \\
 ReAct~\cite{sun2021react}  & 26.45	& 94.95 & 62.27 & 84.47  \\
\textbf{DICE} (ours) & \multicolumn{1}{c}{\textbf{20.83}$^{\pm{1.58}}$} & \textbf{95.24}$^{\pm{0.24}}$ & \multicolumn{1}{c}{\textbf{49.72}$^{\pm{1.69}}$} & \multicolumn{1}{c}{\textbf{87.23}$^{\pm{0.73}}$} \\ \bottomrule  
\end{tabular}}
\label{tab:cifar-results}
\end{table}

\noindent \textbf{ID classification accuracy} Given the \emph{post hoc} nature of DICE, once the input image is marked as ID, one can always use the original fc layer, {which is guaranteed to give identical classification accuracy}. This incurs minimal overhead and results in optimal performance for both classification and OOD detection. We also measure the classification accuracy under 
different sparsification parameter $p$. Due to the space limit, the full results are available in Table~\ref{tab:k-ablation} in Appendix.

\subsection{Evaluation on ImageNet}
\label{sec:imagenet}

\noindent \textbf{Dataset} We then evaluate DICE on a large-scale ImageNet classification model. Following MOS~\cite{huang2021mos}, we use four OOD test datasets from (subsets of) \texttt{Places365}~\cite{zhou2017places}, \texttt{Textures}~\cite{cimpoi2014describing}, \texttt{iNaturalist}~\cite{inat}, and \texttt{SUN}~\cite{sun} with non-overlapping categories \emph{w.r.t.} ImageNet. The evaluations span a diverse range of domains including fine-grained images, scene images, and textural images. OOD detection for the ImageNet model is more challenging due to both a larger feature space ($m=2,048$) as well as a larger label space $(C=1,000)$. In particular, the large-scale evaluation can be relevant to real-world applications, where the deployed models often operate on images that have high resolution and contain many class labels. Moreover, as the number of feature dimensions increases, noisy signals may increase accordingly, which can make OOD detection more challenging.  \\

\noindent \textbf{Experimental details} 
We use a pre-trained ResNet-50 model~\cite{he2016identity} for ImageNet-1k provided by Pytorch.
At test time, all images are resized to 224 $\times$ 224. We use the entire training dataset to estimate the contribution matrix and masking matrix $\*M$. We use the validation strategy in Appendix~\ref{app:val} to select $p$.
The hardware used for experiments is specified in Appendix~\ref{sec:reproduce}. \\

\begin{table*}[t]

\caption[]{\small \textbf{Main results.} Comparison with competitive \emph{post hoc} out-of-distribution detection methods. All methods are based on a discriminative model trained on {ImageNet}. $\uparrow$ indicates larger values are better and $\downarrow$ indicates smaller values are better. All values are percentages. \textbf{Bold} numbers are superior results. 
}
\label{tab:main-results}
\scalebox{0.79}{
\begin{tabular}{lllllllllll}
    \toprule
  \multicolumn{1}{c}{\multirow{4}{*}{\textbf{Methods}}} & \multicolumn{8}{c}{\textbf{OOD Datasets}} & \multicolumn{2}{c}{\multirow{2}{*}{\textbf{Average}}} \\ \cline{2-9}
 \multicolumn{1}{c}{} & \multicolumn{2}{c}{\textbf{iNaturalist}} & \multicolumn{2}{c}{\textbf{SUN}} & \multicolumn{2}{c}{\textbf{Places}} & \multicolumn{2}{c}{\textbf{Textures}} & & \\
 \multicolumn{1}{c}{} & FPR95 & AUROC & FPR95 & AUROC & FPR95 & AUROC & FPR95 & AUROC & FPR95 & AUROC \\
 \multicolumn{1}{l}{} & \multicolumn{1}{l}{$\downarrow$} & \multicolumn{1}{l}{$\uparrow$} & \multicolumn{1}{l}{$\downarrow$} & \multicolumn{1}{l}{$\uparrow$} & \multicolumn{1}{l}{$\downarrow$} & \multicolumn{1}{l}{$\uparrow$} & \multicolumn{1}{l}{$\downarrow$} & \multicolumn{1}{l}{$\uparrow$} & \multicolumn{1}{l}{$\downarrow$} & \multicolumn{1}{l}{$\uparrow$} \\ \hline

MSP ~\cite{Kevin} & 54.99 & 87.74 & 70.83 & 80.86 & 73.99 & 79.76 & 68.00 & 79.61 & 66.95 & 81.99 \\
ODIN ~\cite{liang2018enhancing} & 47.66 & 89.66 & 60.15 & 84.59 & 67.89 & 81.78 & 50.23 & 85.62 & 56.48 & 85.41 \\
GODIN ~\cite{godin2020CVPR} & 61.91 & 85.40 & 60.83 & 85.60	 & 63.70 & 83.81 & 77.85 & 73.27 & 66.07 & 82.02 \\
Mahalanobis \cite{lee2018simple} & 97.00 & 52.65 & 98.50 & 42.41 & 98.40 & 41.79 & 55.80 & 85.01 & 87.43 & 55.47 \\
Energy ~\cite{liu2020energy} & 55.72 & 89.95 & 59.26 & 85.89 & 64.92 & 82.86 & 53.72 & 85.99 & 58.41 & 86.17 \\

ReAct~\cite{sun2021react} & 20.38 & 96.22 & \textbf{24.20} & \textbf{94.20} & \textbf{33.85} & \textbf{91.58} & 47.30 & 89.80 & 31.43 & 92.95 \\ \midrule
\textbf{DICE} (ours) & 25.63 & 94.49 & 35.15 & 90.83 & 46.49 & 87.48 & 31.72 & 90.30 & 34.75 & 90.77 \\

\textbf{DICE + ReAct} (ours) & \textbf{18.64} & \textbf{96.24} & 25.45 & 93.94 & 36.86 & 90.67 & \textbf{28.07} & \textbf{92.74} & \textbf{27.25} & \textbf{93.40}\\
\bottomrule
\end{tabular}
}

\end{table*}

\noindent \textbf{Comparison with baselines} In Table~\ref{tab:main-results}, we compare \methodAbbr with competitive post hoc OOD detection methods. We report performance for each OOD test dataset, as well as the average of the four. 
We first contrast DICE with energy score~\cite{liu2020energy}, which allows us to see the direct benefit of using sparsification under the same scoring function. DICE reduces
the FPR95 drastically from 58.41\% to 34.75\%, a \textbf{23.66}\% improvement using sparsification.  
Second, we contrast with a recent method ReAct~\cite{sun2021react}, which demonstrates strong performance on this challenging task using activation truncation. 
With the truncated activation proposed in ReAct~\cite{sun2021react}, we show that \methodAbbr can further reduce the FPR95 by {5.78}\% with weight sparsification. 
Since the comparison is conducted on the same scoring function and feature activation, the performance improvement from ReAct to DICE+ReAct precisely highlights the benefit of using weight sparsification as opposed to the full weights.
Lastly, Mahalanobis displays limiting performance on ImageNet, while being computationally expensive due to estimating the inverse of the covariance matrix. In contrast, \methodAbbr is easy to use in practice, and can be implemented through simple {post hoc} weight masking. %

\section{Discussion and Ablations}
\label{sec:sparsification}
\begin{figure}[t]
	\begin{center}
		\includegraphics[width=0.7\linewidth]{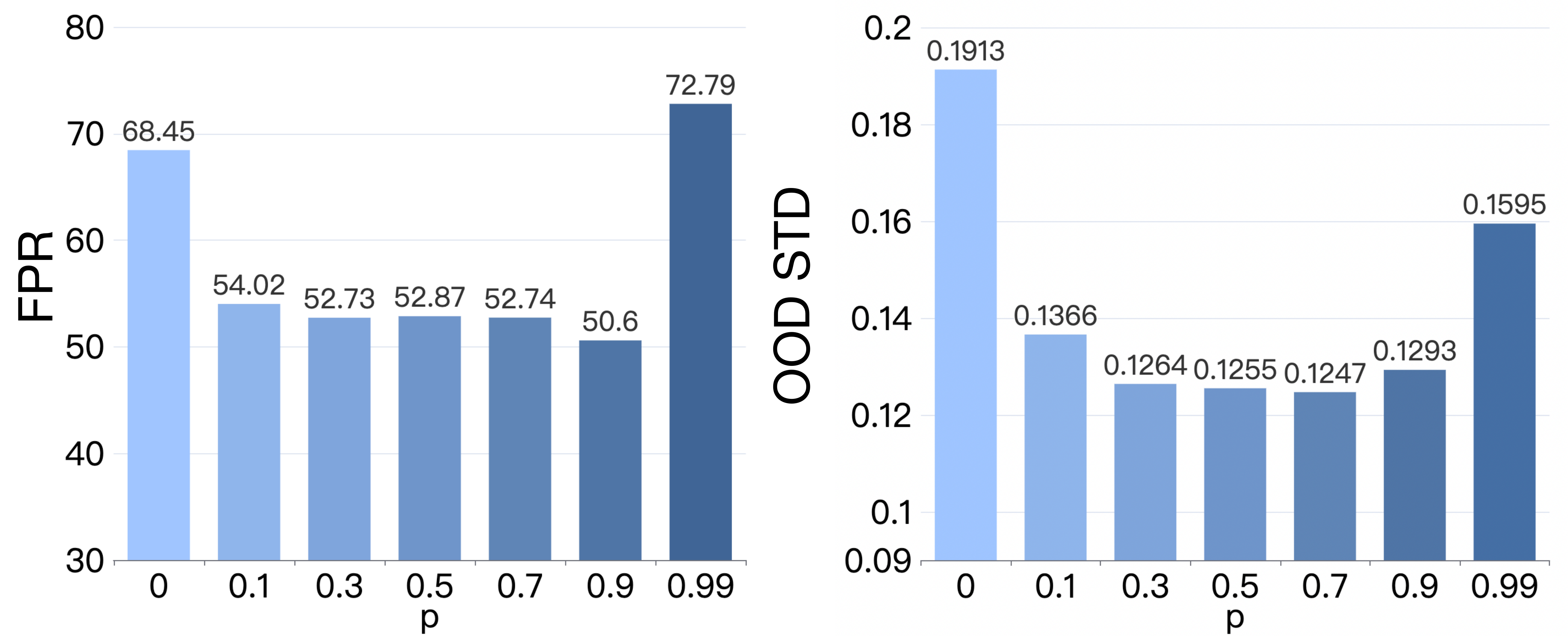}
	\end{center}
	\vspace{-0.3cm}
	\caption{\small Effect of varying sparsity parameter $p$ during inference time. Model is trained on CIFAR-100 using DenseNet101~\cite{huang2017densely}. 
	}
	\vspace{-0.4cm}
	\label{fig:sparsity}
\end{figure}

\noindent \textbf{Ablation on sparsity parameter $p$} 
We now characterize  the effect of sparsity parameter $p$.  In Figure~\ref{fig:sparsity}, we summarize the OOD detection performance for DenseNet trained on CIFAR-100, where we vary $p=\{0.1, 0.3, 0.5, 0.7, 0.9, 0.99\}$.  
Interestingly, we observe the performance improves  with mild sparsity parameter $p$. A significant improvement can be observed from $p=0$ (no sparsity) to $p=0.1$. As we will theoretically later in Section~\ref{sec:theory}, this is because the leftmost part of units being pruned has larger variances for OOD data (gray shade). Units in the middle part have small variances and contributions for both ID and OOD, therefore leading to similar performance as $p$ increases mildly. This ablation confirms that over-parameterization does compromise the OOD detection ability, and \methodAbbr can effectively alleviate the problem. In the extreme case when $p$ is too large (\emph{e.g.}, $p=0.99$), the OOD performance starts to degrade as expected. \\

\noindent \textbf{Effect of variance reduction for output distribution} 
Figure~\ref{fig:teaser} shows that \methodAbbr has an interesting variance reduction effect on the output distribution for OOD data, and at the same time preserves the information for the ID data (CIFAR-10, class ``frog''). The output distribution without any sparsity ($p=0$) appears to have a larger variance, resulting in less separability from ID data (see left of Figure~\ref{fig:teaser}). In contrast, sparsification with \methodAbbr results in a sharper distribution, which benefits OOD detection. 
In Figure~\ref{fig:sparsity}, we also measure the standard deviation of energy score for OOD data (normalized by the mean of ID data's OOD scores in each setting). By way of sparsification, \methodAbbr can reduce the output variance. In Section~\ref{sec:theory}, we formally characterize this and provide a theoretical explanation.

\begin{table}[t]
\caption{\small \textbf{Ablation results.} Effect of different \textit{post hoc} sparsification methods for OOD detection with ImageNet as ID dataset. All sparsification methods are based on the {same OOD scoring function}~\cite{liu2020energy}, with sparsity parameter $p=0.7$. All values are percentages and are averaged over multiple OOD test datasets.}
\label{tab:prune-results}
\centering
\scalebox{1.0}{
        \begin{tabular}{lll}
    \toprule
     \multicolumn{1}{l}{\multirow{1}{*}{\textbf{Method}}} 
    & \multicolumn{1}{l}{\textbf{FPR95}}$\downarrow$ & \multicolumn{1}{l}{\textbf{AUROC}}$\uparrow$  \\
    \midrule
     Weight-Droput   & 76.28 &  76.55  \\
     Unit-Droput   & 83.91  & 64.98 \\
     Weight-Pruning &  52.81 & 87.08 \\
     Unit-Pruning   & 90.80 &  49.15 \\
     {{DICE (Ours)}} & \textbf{34.75} & \textbf{90.77} \\
    \bottomrule
    \end{tabular}
    }
\end{table}

\noindent \textbf{Ablation on  pruning methods} In this ablation, we evaluate OOD detection performance under the most common \emph{post hoc} sparsification methods. Here we primarily consider {post hoc} sparsification strategy which
operates conveniently on a \emph{pre-trained} network, instead of training with sparse regularization or architecture modification. The property is especially desirable for the adoption of OOD detection methods in real-world production environments, where the overhead cost of retraining can be sometimes prohibitive.   Orthogonal to existing works on sparsification, our primary goal is to explore the role of sparsification for improved  OOD detection performance, {rather than establishing a generic sparsification algorithm}. We consider the most common strategies, covering both unit-based and weight-based sparsification methods: (1) {unit dropout}~\cite{Nitish2014dropout} which randomly drops a fraction of units, (2) {unit pruning}~\cite{Hao2017pruneUnit} which
drops units with the smallest $L_2$ norm of the corresponding weight vectors, (3) {weight dropout}~\cite{Wan2013weightdropout} which randomly drops weights in the fully connected layer, and (4) {weight pruning}~\cite{Han2015prune} drops weights with the smallest entries under the $L_1$ norm. %
For consistency, we use the same OOD scoring function and the same sparsity parameter for all. 

Our ablation reveals several important insights shown in Table~\ref{tab:prune-results}. 
First, in contrasting weight dropout vs. DICE, a salient performance gap of {41.53}\% (FPR95) is observed under the same sparsity. This suggests the importance of dropping weights \emph{directedly} rather than \emph{randomly}. Second, \methodAbbr outperforms a popular $L_1$-norm-based pruning method~\cite{Han2015prune} by up to {18.06}\% (FPR95). While it prunes weights with low magnitude, negative weights with large $L_1$-norm can be kept. The negative weights can undesirably corrupt the output with noisy signals (as shown in Figure~\ref{fig:whytopk}). The performance gain of \methodAbbr over~\cite{Han2015prune} attributes to our contribution-directed sparsification, which is better suited for OOD detection.

\begin{table}
    \caption{\small Ablation on different strategies of choosing a subset of units.  Values are FPR95 (averaged over multiple test datasets).}
    \centering
        \begin{tabular}{c|lll}
        \toprule
        Method  & CIFAR-10\textbf{$\downarrow$} & CIFAR-100 \textbf{$\downarrow$}\\
        \midrule
         Bottom-$k$ & 91.87 & 99.70 \\
         (Top+Bottom)-$k$  & 24.25 & 59.93\\
          Random-$k$ & 62.12 & 77.48 \\
         Top-$k$ (\textbf{DICE}) & \textbf{20.83}$^{\pm{1.58}}$ & \textbf{49.72}$^{\pm{1.69}}$ \\
         \bottomrule
        \end{tabular}
    \label{tab:topbot}
\end{table}

\noindent \textbf{Ablation on unit selection}
We have shown that choosing a subset of weights (with \emph{top-k} unit contribution) significantly improves the OOD detection performance. In this ablation, we also analyze those ``lower contribution units'' for OOD detection. Specifically, we consider: (1) \emph{Bottom-k} which only includes $k$ unit contribution with least contribution values, (2) \emph{top+bottom-k} which includes $k$ unit contribution with largest and smallest contribution values, (3) \emph{random-k} which randomly includes $k$ unit contribution and (4) \emph{top-k} which is equivalent to DICE method.  In Table~\ref{tab:topbot}, we show that DICE  outperforms these variants.


\section{Why does DICE improve OOD detection?}
\label{sec:theory}

In this section, we formally explain the mechanism by which reliance on irrelevant units hurts OOD detection and how \methodAbbr effectively mitigates the issue. Our analysis highlights that \methodAbbr reduces the output variance for both ID and OOD data. Below we provide details. \\

\begin{figure}
    \centering
    \includegraphics[width=0.9\textwidth]{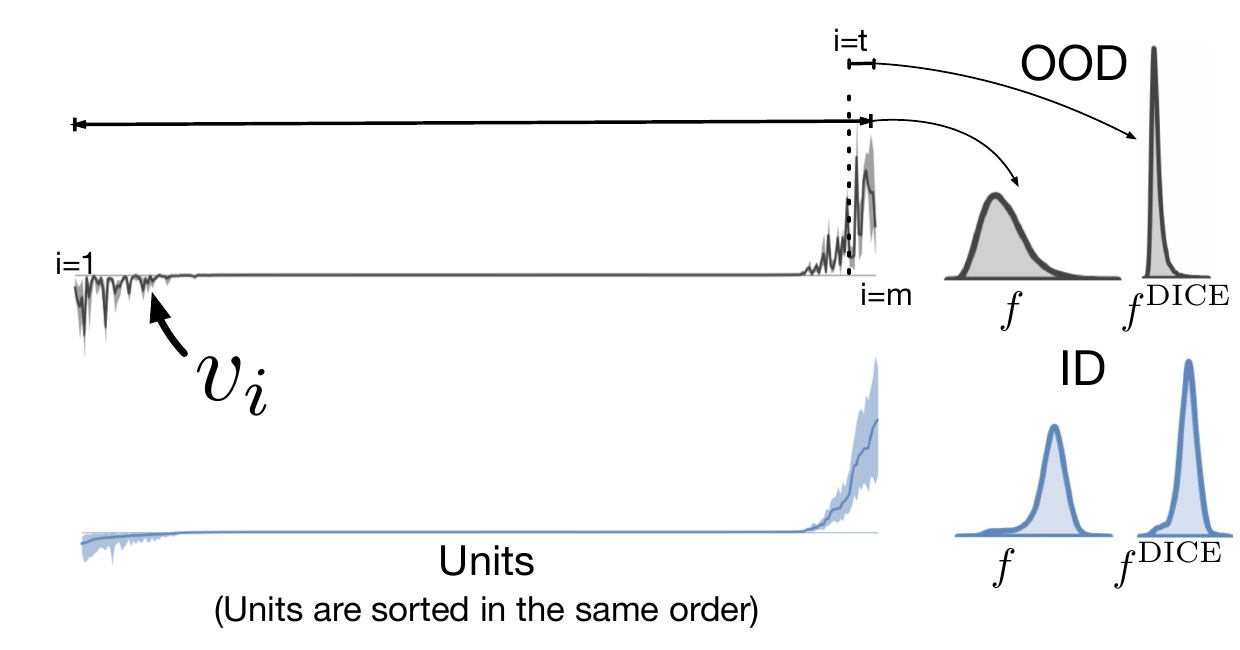}
   \caption{\small Units in the penultimate layer are sorted based on the average contribution to a CIFAR-10 class (``airplane''). OOD data (SVHN) can trigger a non-negligible fraction of units with noisy signals on the CIFAR-10 trained model. }
  \label{fig:theory}
\end{figure}

\noindent \textbf{Setup} For a class $c$, we consider the unit contribution vector $\*v$, the element-wise multiplication between the feature vector $\*h(\*x)$ and corresponding weight vector $\*w$. 
We contrast the two outputs with and without sparsity:
\begin{align*}
f_c= \sum_{i=1}^m v_i ~~~\text{(w.o sparsity)}  , \\
 f_c^\text{DICE}=\sum_{i\in \text{top units}} v_i ~~~ ~~~\text{(w. sparsity)},
\end{align*}
where $f_c$ is the output using the summation of all units' contribution, and $f_c^\text{DICE}$ takes the input from the top units (ranked based on the average contribution on ID data, see bottom of Figure~\ref{fig:theory}). \\

\noindent \textbf{\methodAbbr reduces the output variance} We consider the unit contribution vector for OOD data $\*v \in \mathbb{R}^m$, where each element is a \emph{random variable} $v_i$ with  mean $\mathbb{E}[v_i] = \mu_i$ and variance $\mathrm{Var}[v_i] = \sigma_i^2$.
For simplicity, we assume each component is independent, but our theory can be extended to correlated variables (see Remark 1). 
Importantly, {indices in $\*v$ are sorted based on \emph{the same order} of unit contribution on ID data}. %
By using units on the rightmost side, we now show the key result that DICE reduces the output variance.

\begin{proposition}
\label{prop:sum_gaussian}
Let $v_i$ and $v_j$ be two independent {random variables}. Denote the summation $r=v_i + v_j$, we have $\mathbb{E}[r] = \mathbb{E}[v_i] + \mathbb{E}[v_j]$ and $\mathrm{Var}[r] = \mathrm{Var}[v_i] + \mathrm{Var}[v_j]$.
\end{proposition}

\begin{lemma}
\label{lemma:v3}
When taking the top $m-t$ units, the output variable $f_c^\text{DICE}$ under sparsification has reduced variance:
$$\mathrm{Var} [f_c] - \mathrm{Var} [f_c^\text{DICE}] = \sum_{i=1}^{t} \sigma_i^2 $$
\end{lemma}
\noindent \textit{Proof}. The proof directly follows Proposition 1. \\

\noindent \textbf{Remark 1 (Extension to correlated variables)} We can show in a more general case with correlated variables, the variance reduction is:  %
$$\sum_{i=1}^{t} \sigma_i^2  + 2 \sum_{1\le i < j \le m} \mathrm{Cov}(v_i, v_j) - 2\sum_{t< i <j \le m} \mathrm{Cov}(v_i, v_j), $$
where $\mathrm{Cov}(\cdot ,\cdot )$ is the covariance. Our analysis shows that the covariance matrix primarily consists of 0, which indicates the independence of variables. Moreover, the summation of non-zero entries in the full matrix (i.e., the second term) is greater than that of the submatrix with top units (i.e., the third term), resulting in a larger variance reduction than in Lemma~\ref{lemma:v3}. 
See complete proof in Appendix~\ref{sec:corr}. \\

\noindent \textbf{Remark 2} 
 Energy score is directly compatible with \methodAbbr since \texttt{logsumexp} operation is a smooth approximation~of maximum logit. Our theoretical analysis above shows that DICE reduces the variance of each logit $f_c(\*x)$. 
This means that for detection scores such as energy score, the gap between OOD and ID score will be enlarged after applying DICE, which makes thresholding more capable of separating OOD and ID inputs and benefit OOD detection. \\

\begin{table}[t]
\centering
     \caption{\small Difference between the mean of ID's output and OOD's output. Here we use CIFAR-100 as ID data and {$\Delta$=$\mathbb{E}_\text{in}[\max_c f_c^\text{DICE}]$ - $\mathbb{E}_\text{out}[\max_c f_c^\text{DICE}]$} is averaged over six common OOD benchmark datasets described in Section~\ref{sec:experiments}.}
    \label{tab:mean_shift}
\scalebox{0.9}{
    \centering
    \begin{tabular}{c|cccccc}
         \toprule 
          \textbf{Sparsity} &  $p=0.9$ & $p=0.7$ & $p=0.5$ & $p=0.3$ & $p=0.1$ & $p=0$\\ \midrule
          {$\Delta$ } & 7.92 & 7.28 & 7.99 & 8.04 & 7.36 &  6.67  \\ 
          \bottomrule
 \end{tabular}

    }
\end{table}

\noindent \textbf{Remark 3 (Mean of output)}  Beyond variance, we further show in Table~\ref{tab:mean_shift} the effect of sparsity on the mean of output: $\mathbb{E}_\text{in}[\max_c f_c^\text{DICE}]$ and $\mathbb{E}_\text{out}[\max_c f_c^\text{DICE}]$. The gap between the two directly translates into the OOD score separability. We show that DICE maintains similar (or even enlarges) differences in terms of mean as sparsity $p$ increases. Therefore, DICE overall benefits OOD detection due to both \emph{reduced output variances} and \emph{increased differences of mean}---the combination of both effects leads to stronger separability between ID and OOD. \\

\noindent \textbf{Remark 4 (Variance reduction on ID data)}
Note that we can also show the effect of variance reduction for ID data in a similar way. 
Importantly, \methodAbbr effectively preserves the most important information akin to the ID data, while reducing noisy signals that are harmful to OOD detection. Overall the variance reduction effect on both ID and OOD data leads to stronger separability.


\section{Related Work}
\label{sec:related}
\noindent \textbf{Out-of-distribution detection}  
has attracted growing research attention in recent years. We highlight two major lines of work:

(1) One line of work perform OOD detection by devising scoring functions, including confidence-based methods~\cite{bendale2016towards,Kevin,huang2021mos,liang2018enhancing}, energy-based score~\cite{lin2021mood,liu2020energy,morteza2022provable,wang2021canmulti,sun2021react}, distance-based approaches~\cite{lee2018simple,2021ssd,sun2022knnood,tack2020csi}, gradient-based score~\cite{huang2021importance}, and Bayesian approaches~\cite{gal2016dropout,lakshminarayanan2017simple,maddox2019simple,malinin2018predictive,dpn19nips}. However, {none} of the previous methods considered weight sparsification for OOD detection. 
The closest work to ours is ReAct~\cite{sun2021react}, which proposed truncating the
high activations during test time for OOD detection. While ReAct only considers activation space, DICE examines both the weights and
activation values together---the multiplication of which directly determines the unit contributions to the output. Our work is also related to~\cite{dietterich2022familiarity}, which pointed out that modern OOD detection methods succeed by detecting the existence of familiar features. DICE strengthens the familiarity hypothesis by keeping the dominating weights corresponding to the ``major features''.

(2) A separate line of methods addressed OOD detection by training-time regularization~\cite{bevandic2018discriminative,chen2021atom,geifman2019selectivenet,hein2019relu,hendrycks2018deep,jeong2020ood,katzsamuels2022training,lee2017training,liu2020energy,malinin2018predictive,meinke2019towards,ming2022posterior,mohseni2020self,van2020uncertainty,hongxin2022logitnorm,yang2021semantic}.
For example, models are encouraged to give predictions with uniform distribution~\cite{hendrycks2018deep,lee2017training} or higher energies~\cite{du2022unknown,du2022vos,katzsamuels2022training,liu2020energy,ming2022posterior} for outlier data. The scope of this paper focuses on post hoc methods, which have the advantages of being easy to use and general applicability without modifying the training objective. The latter property is especially desirable for the adoption of OOD detection methods in real-world production environments, when the overhead cost of retraining can be  prohibitive. \\

\noindent \textbf{Pruning and sparsification} 
A great number of effort has been put into improving \emph{post hoc} pruning and training time regularization for deep neural networks~\cite{adadrop2013neurips,Mohammad2016NoiseOut,targetDropout,Han2016deepcomp,Han2015prune,Hao2017pruneUnit,Christos2018l0prune}. Many works obtain a sparse model by training with sparse regularization~\cite{adadrop2013neurips,Mohammad2016NoiseOut,Han2016deepcomp,Christos2018l0prune,sun2019dka} or architecture modification~\cite{targetDropout,Hao2017pruneUnit}, while our work primarily considers \emph{post hoc} sparsification strategy which
operates conveniently on a {pre-trained} network. On this line, two popular Bernoulli dropout techniques include unit dropout and weight dropout~\cite{Nitish2014dropout}. 
\emph{Post hoc} pruning strategies truncate weights with low magnitude~\cite{Han2015prune}, or drop units with low weight norms~\cite{Hao2017pruneUnit}.  %
In ~\cite{wong2021leveraging}, they use a sparse linear layer to help identify spurious correlations and explain misclassifications. 
Orthogonal to existing works, our goal is to improve the OOD detection performance rather than accelerate computation and network debugging. In this paper, we first demonstrate that sparsification can be useful for OOD detection.
An in-depth discussion and comparison of these methods are presented in Section~\ref{sec:sparsification}. \\

\noindent \textbf{Distributional shifts.} Distributional shifts have attracted increasing research interest. It is important to recognize and differentiate various types of distributional shift problems. Literature in OOD detection is commonly concerned about model reliability and detection of semantic shifts, where the OOD inputs have disjoint labels \emph{w.r.t.} ID data and therefore {should not be predicted by the model}. This is different from the OOD generalization task whose goal is to provide accurate predictions on OOD images under the same label space. For example, some works considered covariate shifts in the input space~\cite{koh2021wilds,hendrycks2019benchmarking,ovadia2019can,sun2020test,zhou2021domain}, where the model is expected to generalize to the OOD data.

\section{Conclusion}
This paper provides a simple sparsification strategy termed DICE, which ranks weights based on a contribution measure and then uses the most significant weights to derive the output for OOD detection. We provide both empirical and theoretical insights characterizing and explaining the mechanism by which DICE improves OOD detection. By exploiting the most important weights, DICE provably reduces the output variance for OOD data, resulting in a sharper output distribution and stronger separability from ID data. Extensive experiments show \methodAbbr can significantly improve the performance of OOD detection for over-parameterized networks. We hope our research can raise more attention to the importance of weight sparsification for OOD detection.

\section*{Acknowledgement}

Work was supported by funding from Wisconsin Alumni Research Foundation (WARF). The authors would also like to thank reviewers for the helpful  feedback.

\clearpage

\bibliographystyle{splncs04}
\bibliography{dice}

\begin{thebibliography}{10}
\providecommand{\url}[1]{\texttt{#1}}
\providecommand{\urlprefix}{URL }
\providecommand{\doi}[1]{https://doi.org/#1}

\bibitem{adadrop2013neurips}
Ba, J., Frey, B.: Adaptive dropout for training deep neural networks. In:
  Advances in Neural Information Processing Systems. vol.~26 (2013)

\bibitem{Mohammad2016NoiseOut}
Babaeizadeh, M., Smaragdis, P., Campbell, R.H.: Noiseout: {A} simple way to
  prune neural networks. CoRR  \textbf{abs/1611.06211} (2016)

\bibitem{bendale2016towards}
Bendale, A., Boult, T.E.: Towards open set deep networks. In: Proceedings of
  the IEEE conference on computer vision and pattern recognition. pp.
  1563--1572 (2016)

\bibitem{bevandic2018discriminative}
Bevandi{\'c}, P., Kre{\v{s}}o, I., Or{\v{s}}i{\'c}, M., {\v{S}}egvi{\'c}, S.:
  Discriminative out-of-distribution detection for semantic segmentation. arXiv
  preprint arXiv:1808.07703  (2018)

\bibitem{chen2021atom}
Chen, J., Li, Y., Wu, X., Liang, Y., Jha, S.: Atom: Robustifying
  out-of-distribution detection using outlier mining. In: Proceedings of
  European Conference on Machine Learning and Principles and Practice of
  Knowledge Discovery in Databases (2021)

\bibitem{cimpoi2014describing}
Cimpoi, M., Maji, S., Kokkinos, I., Mohamed, S., Vedaldi, A.: Describing
  textures in the wild. In: Proceedings of the IEEE Conference on Computer
  Vision and Pattern Recognition. pp. 3606--3613 (2014)

\bibitem{dietterich2022familiarity}
Dietterich, T.G., Guyer, A.: The familiarity hypothesis: Explaining the
  behavior of deep open set methods. arXiv preprint arXiv:2203.02486  (2022)

\bibitem{du2022unknown}
Du, X., Wang, X., Gozum, G., Li, Y.: Unknown-aware object detection: Learning
  what you don’t know from videos in the wild. In: Proceedings of the
  IEEE/CVF Conference on Computer Vision and Pattern Recognition (2022)

\bibitem{du2022vos}
Du, X., Wang, Z., Cai, M., Li, Y.: Vos: Learning what you don’t know by
  virtual outlier synthesis. In: Proceedings of the International Conference on
  Learning Representations (2022)

\bibitem{filos2020can}
Filos, A., Tigkas, P., McAllister, R., Rhinehart, N., Levine, S., Gal, Y.: Can
  autonomous vehicles identify, recover from, and adapt to distribution shifts?
  In: Proceedings of the International Conference on Machine Learning. pp.
  3145--3153. PMLR (2020)

\bibitem{gal2016dropout}
Gal, Y., Ghahramani, Z.: Dropout as a bayesian approximation: Representing
  model uncertainty in deep learning. In: Proceedings of the International
  Conference on Machine Learning. pp. 1050--1059 (2016)

\bibitem{geifman2019selectivenet}
Geifman, Y., El-Yaniv, R.: Selectivenet: A deep neural network with an
  integrated reject option. arXiv preprint arXiv:1901.09192  (2019)

\bibitem{targetDropout}
Gomez, A.N., Zhang, I., Kamalakara, S.R., Madaan, D., Swersky, K., Gal, Y.,
  Hinton, G.E.: Learning sparse networks using targeted dropout. arXiv preprint
  arXiv:1905.13678  (2019)

\bibitem{goodfellow2014explaining}
Goodfellow, I.J., Shlens, J., Szegedy, C.: Explaining and harnessing
  adversarial examples. arXiv preprint arXiv:1412.6572  (2014)

\bibitem{Han2016deepcomp}
Han, S., Mao, H., Dally, W.J.: Deep compression: Compressing deep neural
  network with pruning, trained quantization and huffman coding. In:
  Proceedings of the International Conference on Learning Representations
  (2016)

\bibitem{Han2015prune}
Han, S., Pool, J., Tran, J., Dally, W.: Learning both weights and connections
  for efficient neural network. In: Proceedings of the Advances in Neural
  Information Processing Systems. vol.~28, pp. 1135--1143 (2015)

\bibitem{he2016identity}
He, K., Zhang, X., Ren, S., Sun, J.: Identity mappings in deep residual
  networks. In: Proceedings of the European conference on computer vision. pp.
  630--645. Springer (2016)

\bibitem{hein2019relu}
Hein, M., Andriushchenko, M., Bitterwolf, J.: Why relu networks yield
  high-confidence predictions far away from the training data and how to
  mitigate the problem. In: Proceedings of the IEEE Conference on Computer
  Vision and Pattern Recognition. pp. 41--50 (2019)

\bibitem{hendrycks2019benchmarking}
Hendrycks, D., Dietterich, T.: Benchmarking neural network robustness to common
  corruptions and perturbations. arXiv preprint arXiv:1903.12261  (2019)

\bibitem{Kevin}
Hendrycks, D., Gimpel, K.: A baseline for detecting misclassified and
  out-of-distribution examples in neural networks. Proceedings of International
  Conference on Learning Representations  (2017)

\bibitem{hendrycks2018deep}
Hendrycks, D., Mazeika, M., Dietterich, T.: Deep anomaly detection with outlier
  exposure. arXiv preprint arXiv:1812.04606  (2018)

\bibitem{godin2020CVPR}
Hsu, Y.C., Shen, Y., Jin, H., Kira, Z.: Generalized odin: Detecting
  out-of-distribution image without learning from out-of-distribution data. In:
  Proceedings of the IEEE/CVF Conference on Computer Vision and Pattern
  Recognition (2020)

\bibitem{huang2017densely}
Huang, G., Liu, Z., Van Der~Maaten, L., Weinberger, K.Q.: Densely connected
  convolutional networks. In: Proceedings of the IEEE conference on Computer
  Vision and Pattern Recognition. pp. 4700--4708 (2017)

\bibitem{huang2021importance}
Huang, R., Geng, A., Li, Y.: On the importance of gradients for detecting
  distributional shifts in the wild. In: Proceedings of the Advances in Neural
  Information Processing Systems (2021)

\bibitem{huang2021mos}
Huang, R., Li, Y.: Towards scaling out-of-distribution detection for large
  semantic space. In: Proceedings of the IEEE/CVF Conference on Computer Vision
  and Pattern Recognition (2021)

\bibitem{jeong2020ood}
Jeong, T., Kim, H.: Ood-maml: Meta-learning for few-shot out-of-distribution
  detection and classification. In: Proceedings of the Advances in Neural
  Information Processing Systems (2020)

\bibitem{katzsamuels2022training}
Katz-Samuels, J., Nakhleh, J., Nowak, R., Li, Y.: Training ood detectors in
  their natural habitats. In: Proceedings of the International Conference on
  Machine Learning. PMLR (2022)

\bibitem{koh2021wilds}
Koh, P.W., Sagawa, S., Xie, S.M., Zhang, M., Balsubramani, A., Hu, W.,
  Yasunaga, M., Phillips, R.L., Gao, I., Lee, T., et~al.: Wilds: A benchmark of
  in-the-wild distribution shifts. In: Proceedings of the International
  Conference on Machine Learning. pp. 5637--5664. PMLR (2021)

\bibitem{krizhevsky2009learning}
Krizhevsky, A., Hinton, G., et~al.: Learning multiple layers of features from
  tiny images  (2009)

\bibitem{lakshminarayanan2017simple}
Lakshminarayanan, B., Pritzel, A., Blundell, C.: Simple and scalable predictive
  uncertainty estimation using deep ensembles. In: Advances in neural
  information processing systems. pp. 6402--6413 (2017)

\bibitem{lee2017training}
Lee, K., Lee, H., Lee, K., Shin, J.: Training confidence-calibrated classifiers
  for detecting out-of-distribution samples. arXiv preprint arXiv:1711.09325
  (2017)

\bibitem{lee2018simple}
Lee, K., Lee, K., Lee, H., Shin, J.: A simple unified framework for detecting
  out-of-distribution samples and adversarial attacks. In: Advances in Neural
  Information Processing Systems. pp. 7167--7177 (2018)

\bibitem{Hao2017pruneUnit}
Li, H., Kadav, A., Durdanovic, I., Samet, H., Graf, H.P.: Pruning filters for
  efficient convnets. In: Proceedings of International Conference on Learning
  Representations (2017)

\bibitem{liang2018enhancing}
Liang, S., Li, Y., Srikant, R.: Enhancing the reliability of
  out-of-distribution image detection in neural networks. In: Proceedings of
  International Conference on Learning Representations (2018)

\bibitem{lin2021mood}
Lin, Z., Roy, S.D., Li, Y.: Mood: Multi-level out-of-distribution detection.
  In: Proceedings of the IEEE/CVF Conference on Computer Vision and Pattern
  Recognition. pp. 15313--15323 (June 2021)

\bibitem{liu2020energy}
Liu, W., Wang, X., Owens, J., Li, Y.: Energy-based out-of-distribution
  detection. In: Proceedings of the Advances in Neural Information Processing
  Systems (2020)

\bibitem{Christos2018l0prune}
Louizos, C., Welling, M., Kingma, D.P.: Learning sparse neural networks through
  $l_0$ regularization. In: International Conference on Learning
  Representations (2018)

\bibitem{maddox2019simple}
Maddox, W.J., Izmailov, P., Garipov, T., Vetrov, D.P., Wilson, A.G.: A simple
  baseline for bayesian uncertainty in deep learning. Advances in Neural
  Information Processing Systems  \textbf{32},  13153--13164 (2019)

\bibitem{malinin2018predictive}
Malinin, A., Gales, M.: Predictive uncertainty estimation via prior networks.
  In: Advances in Neural Information Processing Systems. pp. 7047--7058 (2018)

\bibitem{dpn19nips}
Malinin, A., Gales, M.: Reverse kl-divergence training of prior networks:
  Improved uncertainty and adversarial robustness. In: Advances in Neural
  Information Processing Systems (2019)

\bibitem{meinke2019towards}
Meinke, A., Hein, M.: Towards neural networks that provably know when they
  don't know. arXiv preprint arXiv:1909.12180  (2019)

\bibitem{ming2022posterior}
Ming, Y., Fan, Y., Li, Y.: Poem: Out-of-distribution detection with posterior
  sampling. In: Proceedings of the International Conference on Machine
  Learning. PMLR (2022)

\bibitem{mohseni2020self}
Mohseni, S., Pitale, M., Yadawa, J., Wang, Z.: Self-supervised learning for
  generalizable out-of-distribution detection. In: AAAI. pp. 5216--5223 (2020)

\bibitem{morteza2022provable}
Morteza, P., Li, Y.: Provable guarantees for understanding out-of-distribution
  detection. Proceedings of the AAAI Conference on Artificial Intelligence
  (2022)

\bibitem{netzer2011reading}
Netzer, Y., Wang, T., Coates, A., Bissacco, A., Wu, B., Ng, A.Y.: Reading
  digits in natural images with unsupervised feature learning  (2011)

\bibitem{nguyen2015deep}
Nguyen, A., Yosinski, J., Clune, J.: Deep neural networks are easily fooled:
  High confidence predictions for unrecognizable images. In: Proceedings of the
  IEEE conference on computer vision and pattern recognition. pp. 427--436
  (2015)

\bibitem{ovadia2019can}
Ovadia, Y., Fertig, E., Ren, J., Nado, Z., Sculley, D., Nowozin, S., Dillon,
  J., Lakshminarayanan, B., Snoek, J.: Can you trust your model's uncertainty?
  evaluating predictive uncertainty under dataset shift. In: Proceedings of the
  Advances in Neural Information Processing Systems. vol.~32, pp. 13991--14002
  (2019)

\bibitem{roy2021does}
Roy, A.G., Ren, J., Azizi, S., Loh, A., Natarajan, V., Mustafa, B., Pawlowski,
  N., Freyberg, J., Liu, Y., Beaver, Z., et~al.: Does your dermatology
  classifier know what it doesn't know? detecting the long-tail of unseen
  conditions. arXiv preprint arXiv:2104.03829  (2021)

\bibitem{2021ssd}
Sehwag, V., Chiang, M., Mittal, P.: Ssd: A unified framework for
  self-supervised outlier detection. In: International Conference on Learning
  Representations (2021)

\bibitem{Nitish2014dropout}
Srivastava, N., Hinton, G., Krizhevsky, A., Sutskever, I., Salakhutdinov, R.:
  Dropout: A simple way to prevent neural networks from overfitting. In:
  Journal of Machine Learning Research. vol.~15, pp. 1929--1958 (2014)

\bibitem{sun2021react}
Sun, Y., Guo, C., Li, Y.: React: Out-of-distribution detection with rectified
  activations. In: Advances in Neural Information Processing Systems (2021)

\bibitem{sun2022knnood}
Sun, Y., Ming, Y., Zhu, X., Li, Y.: Out-of-distribution detection with deep
  nearest neighbors. In: Proceedings of the International Conference on Machine
  Learning (2022)

\bibitem{sun2019dka}
Sun, Y., Ravi, S., Singh, V.: Adaptive activation thresholding: Dynamic routing
  type behavior for interpretability in convolutional neural networks. In:
  Proceedings of the International Conference on Computer Vision (2019)

\bibitem{sun2020test}
Sun, Y., Wang, X., Liu, Z., Miller, J., Efros, A., Hardt, M.: Test-time
  training with self-supervision for generalization under distribution shifts.
  In: Proceedings of the International Conference on Machine Learning. pp.
  9229--9248. PMLR (2020)

\bibitem{tack2020csi}
Tack, J., Mo, S., Jeong, J., Shin, J.: Csi: Novelty detection via contrastive
  learning on distributionally shifted instances. In: Advances in Neural
  Information Processing Systems (2020)

\bibitem{van2020uncertainty}
Van~Amersfoort, J., Smith, L., Teh, Y.W., Gal, Y.: Uncertainty estimation using
  a single deep deterministic neural network. In: Proceedings of the
  International Conference on Machine Learning (2020)

\bibitem{inat}
Van~Horn, G., Mac~Aodha, O., Song, Y., Cui, Y., Sun, C., Shepard, A., Adam, H.,
  Perona, P., Belongie, S.: The inaturalist species classification and
  detection dataset. In: Proceedings of the IEEE conference on Computer Vision
  and Pattern Recognition. pp. 8769--8778 (2018)

\bibitem{Wan2013weightdropout}
Wan, L., Zeiler, M.D., Zhang, S., LeCun, Y., Fergus, R.: Regularization of
  neural networks using dropconnect. In: Proceedings of the International
  Conference on Machine Learning. vol.~28, pp. 1058--1066 (2013)

\bibitem{wang2021canmulti}
Wang, H., Liu, W., Bocchieri, A., Li, Y.: Can multi-label classification
  networks know what they don't know? Proceedings of the Advances in Neural
  Information Processing Systems  (2021)

\bibitem{wang2017chestx}
Wang, X., Peng, Y., Lu, L., Lu, Z., Bagheri, M., Summers, R.M.: Chestx-ray8:
  Hospital-scale chest x-ray database and benchmarks on weakly-supervised
  classification and localization of common thorax diseases. In: Proceedings of
  the IEEE conference on computer vision and pattern recognition. pp.
  2097--2106 (2017)

\bibitem{hongxin2022logitnorm}
Wei, H., Xie, R., Cheng, H., Feng, L., An, B., Li, Y.: Mitigating neural
  network overconfidence with logit normalization. Proceedings of the
  International Conference on Machine Learning  (2022)

\bibitem{wong2021leveraging}
Wong, E., Santurkar, S., Madry, A.: Leveraging sparse linear layers for
  debuggable deep networks. In: Proceedings of the International Conference on
  Machine Learning. pp. 11205--11216. PMLR (2021)

\bibitem{sun}
Xiao, J., Hays, J., Ehinger, K.A., Oliva, A., Torralba, A.: Sun database:
  Large-scale scene recognition from abbey to zoo. In: Proceedings of the IEEE
  conference on Computer Vision and Pattern Recognition. pp. 3485--3492. IEEE
  Computer Society (2010)

\bibitem{xu2015turkergaze}
Xu, P., Ehinger, K.A., Zhang, Y., Finkelstein, A., Kulkarni, S.R., Xiao, J.:
  Turkergaze: Crowdsourcing saliency with webcam based eye tracking. arXiv
  preprint arXiv:1504.06755  (2015)

\bibitem{yang2021semantic}
Yang, J., Wang, H., Feng, L., Yan, X., Zheng, H., Zhang, W., Liu, Z.:
  Semantically coherent out-of-distribution detection. In: Proceedings of the
  IEEE International Conference on Computer Vision. pp. 8301--8309 (October
  2021)

\bibitem{yu2015lsun}
Yu, F., Seff, A., Zhang, Y., Song, S., Funkhouser, T., Xiao, J.: Lsun:
  Construction of a large-scale image dataset using deep learning with humans
  in the loop. arXiv preprint arXiv:1506.03365  (2015)

\bibitem{zhang2016understanding}
Zhang, C., Bengio, S., Hardt, M., Recht, B., Vinyals, O.: Understanding deep
  learning requires rethinking generalization. In: Proceedings of International
  Conference on Learning Representations

\bibitem{zhou2017places}
Zhou, B., Lapedriza, A., Khosla, A., Oliva, A., Torralba, A.: Places: A 10
  million image database for scene recognition. In: IEEE Transactions on
  Pattern Analysis and Machine Intelligence. vol.~40, pp. 1452--1464. IEEE
  (2017)

\bibitem{zhou2021domain}
Zhou, K., Liu, Z., Qiao, Y., Xiang, T., Loy, C.C.: Domain generalization: A
  survey  (2021)

\end{thebibliography}


\appendix
\clearpage
\newpage

\onecolumn

\section{Reproducibility}
    \label{sec:reproduce}
 Authors of the paper recognize the importance and value of reproducible research. 
 We summarize our efforts below to facilitate reproducible results: 
 \begin{enumerate}
     \item \textbf{Dataset.} We use {publicly available} datasets, which are described in detail in   \emph{Section~\ref{sec:imagenet}} and \emph{Section~\ref{sec:common_benchmark}}. 
     \item \textbf{Assumption and proof.} The complete proof of our theoretical contribution is provided in \emph{Appendix~\ref{sec:corr}}, which supports our theoretical claims made in \emph{Section~\ref{sec:theory}}.
     \item \textbf{Baselines.} The description and hyperparameters of baseline methods are specified in \emph{Appendix~\ref{sec:baseline}}.
     \item \textbf{Model.} Our main results on ImageNet are based on ResNet50~\cite{he2016identity} provided by Pytorch. Due to the post hoc nature of our method, this allows the research community to reproduce our numbers provided with the same model and evaluation datasets. 
     \item  \textbf{Implementation.} The simplicity of the DICE eases the reproducibility, as it only requires a few lines of code modification in the PyTorch model specification. Specifically, one can replace the weight matrix in the penultimate layer of deep networks using the following code:
\begin{lstlisting}[language=Python]
        threshold = numpy.percentile(V, p)
        M = V > threshold
        W_new = W * M
\end{lstlisting}
    \item  \textbf{Open Source.}  The codebase and the dataset is available in  \url{https://github.com/deeplearning-wisc/dice.git}.
    \item \textbf{Hardware}: We conduct all the experiments on NVIDIA GeForce RTX 2080Ti GPUs.
 \end{enumerate}

\section{Details of Baselines}
\label{sec:baseline}

For the reader's convenience, we summarize in detail a few common techniques for defining OOD scores that measure the degree of ID-ness on a given input.
By convention, a higher (lower) score is indicative of being in-distribution (out-of-distribution).

\noindent \textbf{MSP~\cite{Kevin}} This method proposes to use the maximum softmax score as the OOD score.

\noindent \textbf{ODIN~\cite{liang2018enhancing}} This method improves OOD detection with temperature scaling and input perturbation. In all experiments, we set the temperature scaling parameter $T = 1000$. 
For ImageNet, we found the input perturbation does not further improve the OOD detection performance and hence we set $\epsilon=0$. 
Following the setting in~\cite{liang2018enhancing}, we set $\epsilon$ to be 0.004 for CIFAR-10 and CIFAR-100.

\noindent \textbf{Mahalanobis~\cite{lee2018simple}}  This method uses multivariate Gaussian distributions to model class-conditional distributions of softmax neural classifiers and uses Mahalanobis distance-based scores for OOD detection. We use 500 examples randomly selected from ID datasets and an auxiliary tuning dataset to train the logistic regression model and tune the perturbation magnitude $\epsilon$. The tuning dataset consists of adversarial examples generated by FGSM~\cite{goodfellow2014explaining} with a perturbation size of 0.05.\@
The selected $\epsilon$'s are 0.001, 0.0, and 0.0 for ImageNet-1k, CIFAR-10, and CIFAR-100, respectively.

\noindent \textbf{Generalized ODIN~\cite{godin2020CVPR}}  This method proposes a specialized network to learn temperature scaling and a novel strategy to choose perturbation magnitude, in order to replace manually-set hyperparameters. Our training configurations strictly follow the original paper, where we train the DeConf-C network for 200 epochs without applying the weight decay in the final layer of the Deconf classifier (notated as $h_i(x)$ in \cite{godin2020CVPR}). The other settings such as learning rate, momentum and training batch size are the same as ours. Note that  G-ODIN has a slight advantage due to a longer training time than ours (100 epochs). We choose the best perturbation magnitude $\epsilon$ by maximizing the confidence scores on 1,000 examples randomly selected from ID datasets. The selected $\epsilon$ value is 0.02 for all (ImageNet-1k, CIFAR-10, and CIFAR-100).

\noindent \textbf{Energy~\cite{liu2020energy}}  This method proposes using energy score for OOD detection. The energy function  maps the logit outputs to a scalar $E(\*x; f) \in \mathbb{R}$, which is relatively lower for ID data.\@
Note that \cite{liu2020energy} used the \emph{negative energy score} for OOD detection, in order to align with the convention that $S(\*x)$ is higher (lower) for ID (OOD) data. Energy score does not require hyperparameter tuning. 

\noindent \textbf{ReAct~\cite{sun2021react}} This method also uses energy score for OOD detection. It further truncates the internal activations of neural networks, which provides more distinctive feature patterns for OOD distributions. The truncation threshold is set with the validation strategy in~\cite{sun2021react}.

\section{Validation Strategy}
\label{app:val}
We use a validation set of \texttt{Gaussian noise} images, which are generated by sampling from $\mathcal{N}(0,1)$ for each pixel location.
The optimal $p$ is selected from $\{0.1,0.3, 0.5,0.7,0.9,0.99\}$, which is $0.9$ for CIFAR-10/100 and $0.7$ for ImageNet.
We also show in Figure~\ref{fig:sparsity} using Gaussian can already find the near-optimal one averaged over all OOD test datasets considered. 

\section{More results on the effect of Sparsity Parameter $p$ } 
\label{sec:ood-sparsep}
We characterize  the effect of sparsity parameter $p$ on other ID datasets.  In Table~\ref{tab:k-ablation}, we summarize the OOD detection performance and classification performance for DenseNet trained on CIFAR-10 and ImageNet, where we vary $p=\{0.1, 0.3, 0.5, 0.7, 0.9, 0.99\}$. A similar trend is observed on CIFAR-100 as discussed in the main paper.

\begin{table*}[htb] 
\caption[]{\small Effect of varying sparsity parameter $p$. Results are averaged on the test datasets described in Section~\ref{sec:experiments}.}
\label{tab:k-ablation}
\centering
\footnotesize
\scalebox{0.8}{
\begin{tabular}{l|lll|lll}
\toprule
\multirow{2}{*}{\textbf{Sparsity}}  & \multicolumn{3}{c|}{\textbf{CIFAR-10}} & \multicolumn{3}{c}{\textbf{ImageNet}} \\
  & \textbf{FPR95} $\downarrow$ & \textbf{AUROC} $\uparrow$ & \textbf{Acc.} $\uparrow$ & \textbf{FPR95} $\downarrow$ & \textbf{AUROC} $\uparrow$ & \textbf{Acc.} $\uparrow$ \\ \midrule
$p=0.99$ & 57.57 & 84.29 & 60.81 & 75.79 &  66.07 & 63.28\\
$p=0.9$ & 21.76 & 94.91 & 94.38 & 40.10 & 89.09  & 73.36 \\
$p=0.7$ & 21.76 & 94.91 & 94.35 & 34.75 & 90.77  & 73.82 \\
$p=0.5$ & 21.76 & 94.91 & 94.35  & 34.58 & 90.80 & 73.80\\
$p=0.3$ & 21.75 & 94.91 & 94.35 & 34.70 & 90.69 & 73.57 \\
$p=0.1$ & 21.92 & 94.90 & 94.33 & 40.25 & 89.44 &  73.38 \\ \midrule
$p=0$  & 26.55 & 94.57 & 94.50 & 58.41 & 86.17 & 75.20\\ \bottomrule
\end{tabular}}

\end{table*}

\section{Variance Reduction with Correlated Variables}
\label{sec:corr}

\noindent \textbf{Extension of Lemma 2.} We can show variance reduction in a more general case with correlated variables. The variance of output $f_c$ without sparsification is:
$$\mathrm{Var} [f_c] = \sum_{i=1}^{m} \sigma_i^2  + 2 \sum_{1\le i < j \le m} \mathrm{Cov}(v_i, v_j),$$
where $\mathrm{Cov}(\cdot ,\cdot )$ is the covariance. The expression states that the variance is the sum of the diagonal of the covariance matrix plus two times the sum of its upper triangular elements.

Similarly, the variance of output \emph{with} directed sparsification (by taking the top units) is:
$$\mathrm{Var} [f_c^\text{DICE}]= \sum_{i=t+1}^{m} \sigma_i^2  + 2 \sum_{t< i < j \le m} \mathrm{Cov}(v_i, v_j).$$
Therefore, the variance reduction is given by: %
$$\sum_{i=1}^{t} \sigma_i^2  + 2 \sum_{1\le i < j \le m} \mathrm{Cov}(v_i, v_j) - 2\sum_{t< i <j \le m} \mathrm{Cov}(v_i, v_j), $$

\begin{figure*}[ht]
	\begin{center}
		\includegraphics[width=0.7\linewidth]{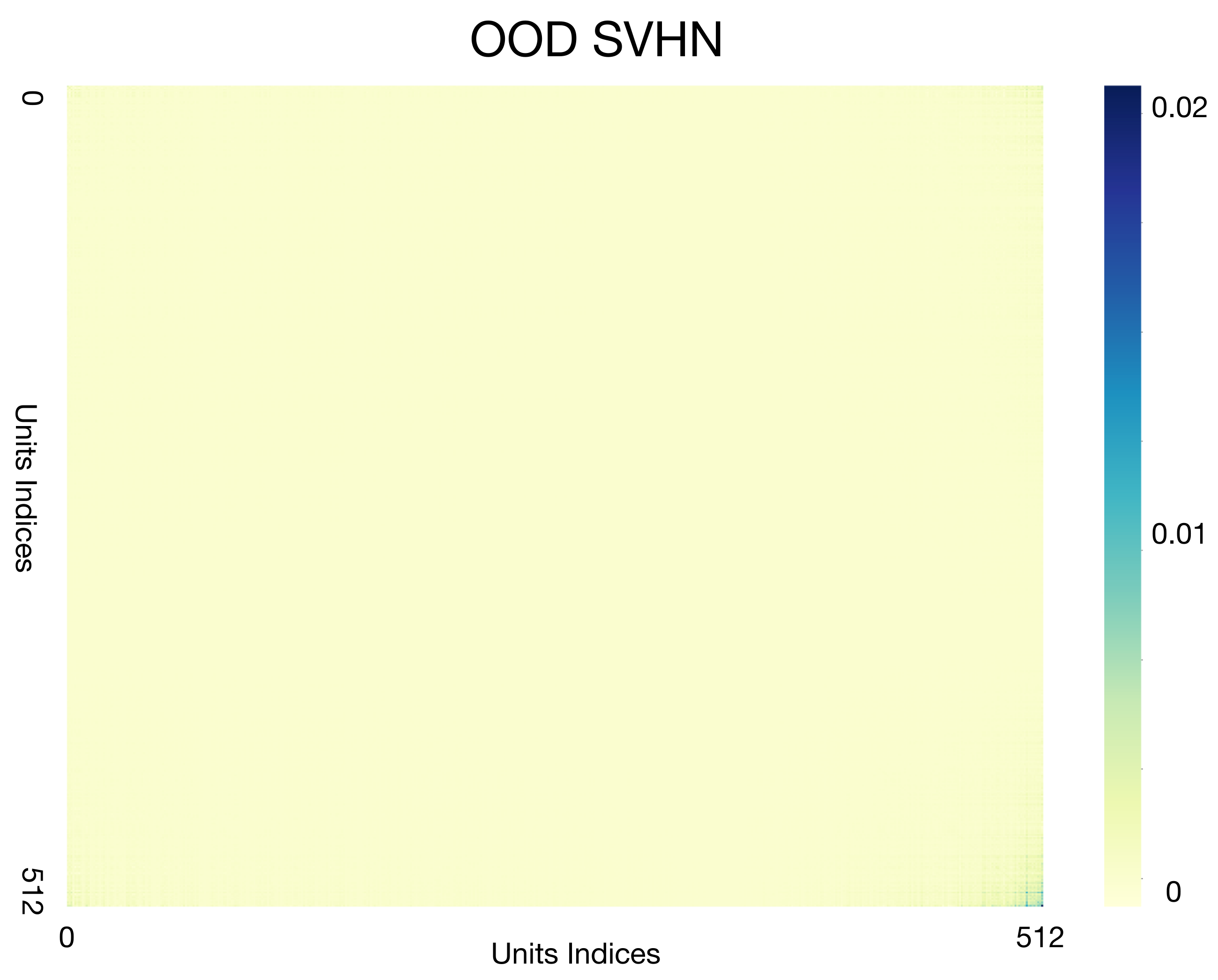}
	\end{center}
	\caption{Covariance matrix of unit contribution estimated on the OOD dataset SVHN. Model is trained on ID dataset CIFAR-10. The unit indices are sorted from low to high, based on the expectation value of ID's unit contribution (airplane class, same as in Figure~\ref{fig:whytopk}). The matrix primarily consists of elements with 0 value. }
	\label{fig:covmat}
\end{figure*}

We show in Fig.~\ref{fig:covmat} that the covariance matrix of unit contribution $v$ primarily consists of elements of 0, which indicates the independence of variables by large. The covariance matrix is estimated on the CIFAR-10 model with DenseNet-101, which is consistent with our main results in Table~\ref{tab:main-results}. %

Moreover, the summation of non-zero entries in the full matrix (i.e., the second term) is greater than that of the submatrix with top units (i.e., the third term), resulting in a larger variance reduction than in Lemma~\ref{lemma:v3}. In the case of OOD data (SVHN), we empirically measure the  variance reduction, where
$\sum_{i=1}^{t} \sigma_i^2  + 2 \sum_{1\le i < j \le m} \mathrm{Cov}(v_i, v_j)$ equals to \textbf{6.8} and $2\sum_{t< i <j \le m} \mathrm{Cov}(v_i, v_j)$ equals to \textbf{2.2}. Therefore, \methodAbbr leads to a significant variance reduction effect. 

\section{Effect of DICE on MSP}
\label{sec:dice_msp}
Our theory shows the variance reduction effect directly in the logit output space, which is more compatible with the energy score. As a further investigation in Table~\ref{tab:dice_msp}, we find empirically that using DICE for MSP can improve the performance for MSP though it does not yield better performance than our main results. 

\begin{table*}[ht]
\caption[]{\small Effect of applying \methodAbbr with MSP on DenseNet101 pretrained on CIFAR-10. The number is reported in FPR95. }
\centering
\scalebox{0.85}{
\begin{tabular}{llllllll} \toprule
Method   & SVHN  & LSUN-c & LSUN-r & iSUN  & Texture & places365 & Average \\ \midrule
MSP~\cite{Kevin}      & 48.25 &	33.80 &	42.37 &	41.42 &	63.99 &	62.57  & 48.73   \\
DICE+MSP & \textbf{45.94} & \textbf{24.36}  & \textbf{35.68}  & \textbf{34.60}  & \textbf{62.06}   & \textbf{59.40} & \textbf{43.67} \\ \bottomrule
\end{tabular}}
\label{tab:dice_msp}
\end{table*}

\section{Detailed OOD Detection Performance for CIFAR}
\label{sec:detailed-cifar}
We report the detailed performance for all six test OOD datasets for models trained on CIFAR10 and CIFAR-100 respectively in Table~\ref{tab:detail-results-cifar10}
 and Table~\ref{tab:detail-results-cifar100}.

\begin{sidewaystable}
\caption{\small Detailed results on six common OOD benchmark datasets: \texttt{Textures}~\cite{cimpoi2014describing}, \texttt{SVHN}~\cite{netzer2011reading}, \texttt{Places365}~\cite{zhou2017places}, \texttt{LSUN-Crop}~\cite{yu2015lsun}, \texttt{LSUN-Resize}~\cite{yu2015lsun}, and \texttt{iSUN}~\cite{xu2015turkergaze}. For each ID dataset, we use the same DenseNet pretrained on \textbf{CIFAR-10}. $\uparrow$ indicates larger values are better and $\downarrow$ indicates smaller values are better.}
\scalebox{0.75}{
\begin{tabular}{llllllllllllllll} \toprule
\multirow{3}{*}{\textbf{Method Type}} & \multirow{3}{*}{\textbf{Method}} & \multicolumn{2}{c}{\textbf{SVHN}} & \multicolumn{2}{c}{\textbf{LSUN-c}} & \multicolumn{2}{c}{\textbf{LSUN-r}} & \multicolumn{2}{c}{\textbf{iSUN}} & \multicolumn{2}{c}{\textbf{Textures}} & \multicolumn{2}{c}{\textbf{Places365}} & \multicolumn{2}{c}{\textbf{Average}} \\ \cline{3-16}
 & & \textbf{FPR95} & \textbf{AUROC} & \textbf{FPR95} & \textbf{AUROC} & \textbf{FPR95} & \textbf{AUROC} & \textbf{FPR95} & \textbf{AUROC} & \textbf{FPR95} & \textbf{AUROC} & \textbf{FPR95} & \textbf{AUROC} & \textbf{FPR95} & \textbf{AUROC} \\ 
 & & $\downarrow$ & $\uparrow$ & $\downarrow$ & $\uparrow$ & $\downarrow$ & $\uparrow$ & $\downarrow$ & $\uparrow$ & $\downarrow$ & $\uparrow$ & $\downarrow$ & $\uparrow$ & $\downarrow$ & $\uparrow$ \\ \midrule
\multirow{5}{*}{Non-Sparse} & MSP  & 47.24 & 93.48 & 33.57 & 95.54 & 42.10 & 94.51 & 42.31 & 94.52 & 64.15 & 88.15 & 63.02 & 88.57 & 48.73 & 92.46 \\
 & ODIN  & 25.29 & 94.57 & 4.70 & 98.86 & 3.09 & 99.02 & 3.98 & 98.90 & 57.50 & 82.38 & 52.85 & 88.55 & 24.57 & 93.71 \\
 & GODIN  & 6.68 & 98.32 & 17.58 & 95.09 & 36.56 & 92.09 & 36.44 & 91.75 & 35.18 & 89.24 & 73.06 & 77.18 & 34.25 & 90.61 \\
 & Mahalanobis  & 6.42 & 98.31 & 56.55 & 86.96 & 9.14 & 97.09 & 9.78 & 97.25 & 21.51 & 92.15 & 85.14 & 63.15 & 31.42 & 89.15 \\
 & Energy  & 40.61 & 93.99 & 3.81 & 99.15 & 9.28 & 98.12 & 10.07 & 98.07 & 56.12 & 86.43 & 39.40 & 91.64 & 26.55 & 94.57 \\ 
 & ReAct  & 41.64 & 93.87 & 5.96 & 98.84 & 11.46 & 97.87 & 12.72 & 97.72 & 43.58 & 92.47 & 43.31 & 91.03 & 26.45 & 94.67 \\
  \midrule
\multirow{5}{*}{Sparse} & Unit-Droput  & 89.16 & 60.96 & 72.97 & 81.33 & 87.03 & 68.78 & 87.29 & 68.07 & 88.53 & 60.10 & 94.82 & 59.18 & 86.63 & 66.40 \\
 & Weight-Droput & 81.34 & 80.03 & 21.06 & 96.15 & 54.70 & 90.33 & 58.88 & 89.80 & 83.34 & 73.31 & 73.42 & 81.10 & 62.12 & 85.12 \\
 & Unit-Pruning & 40.56 & 93.99 & 3.81 & 99.15 & 9.28 & 98.12 & 10.07 & 98.07 & 56.1 & 86.43 & 39.47 & 91.64 & 26.55 & 94.57  \\
 & Weight-Pruning & 28.61 & 95.40 & 3.01 & 99.30 & 8.58 & 98.19 & 9.08 & 98.16 & 49.45 & 88.20 & 46.78 & 89.77 & 24.25 & 94.84 \\
 & {DICE} (ours) 
 & 25.99$^{\pm{5.10}}$ & 95.90$^{\pm{1.08}}$ & 0.26$^{\pm{0.11}}$ & 99.92$^{\pm{0.02}}$ & 3.91$^{\pm{0.56}}$ & 99.20$^{\pm{0.15}}$ & 4.36$^{\pm{0.71}}$ & 99.14$^{\pm{0.15}}$ & 41.90$^{\pm{4.41}}$ & 88.18$^{\pm{1.80}}$ & 48.59$^{\pm{1.53}}$ & 89.13$^{\pm{0.31}}$ & 20.83$^{\pm{1.58}}$ & 95.24$^{\pm{0.24}}$ 
\\ \bottomrule
\end{tabular}}

\label{tab:detail-results-cifar10}
\end{sidewaystable}

\begin{sidewaystable}
\caption{\small Detailed results on six common OOD benchmark datasets: \texttt{Textures}~\cite{cimpoi2014describing}, \texttt{SVHN}~\cite{netzer2011reading}, \texttt{Places365}~\cite{zhou2017places}, \texttt{LSUN-Crop}~\cite{yu2015lsun}, \texttt{LSUN-Resize}~\cite{yu2015lsun}, and \texttt{iSUN}~\cite{xu2015turkergaze}. For each ID dataset, we use the same DenseNet pretrained on \textbf{CIFAR-100}. $\uparrow$ indicates larger values are better and $\downarrow$ indicates smaller values are better. }
\scalebox{0.75}{
\begin{tabular}{llllllllllllllll} \toprule
\multirow{3}{*}{\textbf{Method Type}} & \multirow{3}{*}{\textbf{Method}} & \multicolumn{2}{c}{\textbf{SVHN}} & \multicolumn{2}{c}{\textbf{LSUN-c}} & \multicolumn{2}{c}{\textbf{LSUN-r}} & \multicolumn{2}{c}{\textbf{iSUN}} & \multicolumn{2}{c}{\textbf{Textures}} & \multicolumn{2}{c}{\textbf{Places365}} & \multicolumn{2}{c}{\textbf{Average}} \\ \cline{3-16}
 & & \textbf{FPR95} & \textbf{AUROC} & \textbf{FPR95} & \textbf{AUROC} & \textbf{FPR95} & \textbf{AUROC} & \textbf{FPR95} & \textbf{AUROC} & \textbf{FPR95} & \textbf{AUROC} & \textbf{FPR95} & \textbf{AUROC} & \textbf{FPR95} & \textbf{AUROC} \\ 
 & & $\downarrow$ & $\uparrow$ & $\downarrow$ & $\uparrow$ & $\downarrow$ & $\uparrow$ & $\downarrow$ & $\uparrow$ & $\downarrow$ & $\uparrow$ & $\downarrow$ & $\uparrow$ & $\downarrow$ & $\uparrow$ \\ \midrule
\multirow{5}{*}{\textbf{Non-Sparse}} & MSP & 81.70 & 75.40 & 60.49 & 85.60 & 85.24 & 69.18 & 85.99 & 70.17 & 84.79 & 71.48 & 82.55 & 74.31 & 80.13 & 74.36 \\
 & ODIN & 41.35 & 92.65 & 10.54 & 97.93 & 65.22 & 84.22 & 67.05 & 83.84 & 82.34 & 71.48 & 82.32 & 76.84 & 58.14 & 84.49 \\
 & GODIN & 36.74 & 93.51 & 43.15 & 89.55 & 40.31 & 92.61 & 37.41 & 93.05 & 64.26 & 76.72 & 95.33 & 65.97 & 52.87 & 85.24 \\ 
 & Mahalanobis & 22.44 & 95.67 & 68.90 & 86.30 & 23.07 & 94.20 & 31.38 & 93.21 & 62.39 & 79.39 & 92.66 & 61.39 & 55.37 & 82.73 \\
 & Energy & 87.46 & 81.85 & 14.72 & 97.43 & 70.65 & 80.14 & 74.54 & 78.95 & 84.15 & 71.03 & 79.20 & 77.72 & 68.45 & 81.19 \\ 
 & ReAct & 83.81 & 81.41 & 25.55 & 94.92 & 60.08 & 87.88 & 65.27 & 86.55 & 77.78 & 78.95 & 82.65 & 74.04 & 62.27 & 84.47 \\ \midrule
 \multirow{5}{*}{\textbf{Sparse}} & Unit-Droput & 91.43 & 54.71 & 56.24 & 85.25 & 91.06 & 57.79 & 90.88 & 57.90 & 89.59 & 54.57 & 94.15 & 56.15 & 85.56 & 61.06 \\
 & Weight-Droput & 92.97 & 64.39 & 18.96 & 95.62 & 88.67 & 65.48 & 87.12 & 67.82 & 88.45 & 64.38 & 88.69 & 71.87 & 77.48 & 71.59 \\
 & Unit-Pruning & 87.52 & 81.83 & 14.73 & 97.43 & 70.62 & 80.18 & 74.46 & 79.00 & 84.20 & 71.02 & 79.32 & 77.70 & 68.48 & 81.19 \\
 & Weight-Pruning & 77.99 & 84.14 & 5.17 & 99.05 & 59.42 & 87.13 & 61.80 & 86.09 & 72.68 & 73.85 & 82.53 & 75.06 & 59.93 & 84.22 \\
 & {DICE} (ours) 
  & 54.65$^{\pm{4.94}}$ & 88.84$^{\pm{0.39}}$ & 0.93$^{\pm{0.07}}$ & 99.74$^{\pm{0.01}}$ & 49.40$^{\pm{1.99}}$ & 91.04$^{\pm{1.49}}$ & 48.72$^{\pm{1.55}}$ & 90.08$^{\pm{1.36}}$ & 65.04$^{\pm{0.66}}$ & 76.42$^{\pm{0.35}}$ & 79.58$^{\pm{2.34}}$ & 77.26$^{\pm{1.08}}$ & 49.72$^{\pm{1.69}}$ & 87.23$^{\pm{0.73}}$ 
\\ \bottomrule
\end{tabular}}

\label{tab:detail-results-cifar100}
\end{sidewaystable}

\end{document}